\documentclass[]{spie}  

\usepackage{amsmath,amsfonts,amssymb}
\usepackage{graphicx}
\usepackage{float}
\restylefloat{table}
\usepackage[colorlinks=true, allcolors=blue]{hyperref}
\usepackage{caption}
\usepackage{subcaption}
\usepackage{xurl}

\title{Improving Object Detector Training on Synthetic Data by Starting With a Strong Baseline Methodology}

\author[a]{Frank A. Ruis}
\author[a]{Alma M. Liezenga}
\author[a]{Friso G. Heslinga}
\author[a]{Luca Ballan}
\author[a]{Thijs A. Eker}
\author[a]{Richard J. M. den Hollander}
\author[a]{Martin C. van Leeuwen}
\author[a]{Judith Dijk}
\author[a]{Wyke Huizinga}
\affil[a]{TNO, Oude Waalsdorperweg 63, The Hague, The Netherlands}

\authorinfo{Please send further correspondence to Frank Ruis or Alma Liezenga\\Frank A. Ruis: E-mail: frank.ruis@tno.nl, Telephone: +31 (0)6 11 28 04 89 \\  Alma M. Liezenga: E-mail: alma.liezenga@tno.nl, Telephone: +31 (0)6 27 86 46 08}

\begin{document} 
\maketitle

\begin{abstract}
Collecting and annotating real-world data for the development of object detection models is a time-consuming and expensive process. In the military domain in particular, data collection can also be dangerous or infeasible. Training models on synthetic data may provide a solution for cases where access to real-world training data is restricted. However, bridging the reality gap between synthetic and real data remains a challenge. Existing methods usually build on top of baseline Convolutional Neural Network (CNN) models that have been shown to perform well when trained on real data, but have limited ability to perform well when trained on synthetic data. For example, some architectures allow for fine-tuning with the expectation of large quantities of training data and are prone to overfitting on synthetic data. Related work usually ignores various best practices from object detection on real data, e.g. by training on synthetic data from a single environment with relatively little variation. In this paper we propose a methodology for improving the performance of a pre-trained object detector when training on synthetic data. Our approach focuses on extracting the salient information from synthetic data without forgetting useful features learned from pre-training on real images. Based on the state of the art, we incorporate data augmentation methods and a Transformer backbone. Besides reaching relatively strong performance without any specialized synthetic data transfer methods, we show that our methods improve the state of the art on synthetic data trained object detection for the RarePlanes and DGTA-VisDrone datasets, and reach near-perfect performance on an in-house vehicle detection dataset.
\end{abstract}



\section{INTRODUCTION} 
\label{sec:intro}

The development of accurate object detection models remains a challenge, especially when dealing with data scarcity. The proliferation of synthetic datasets, driven by advancements in graphics technology and simulation, has opened new avenues for training object detection models \cite{vanherle2022analysis, Eker2023, rajpura2017object, prakash2019structured}. Synthetic data offers the advantage of scalability, cost-effectiveness, and control over various environmental factors. However, bridging the reality gap, the difference between synthetic or simulated and real(-world) data, remains a significant hurdle. A substantial body of research has addressed the domain gap, focusing on domain transfer methods such as domain randomization \cite{tobin2017domain, hinterstoisser2019annotation, synthdet2020} or domain adaptation \cite{yang2007cross, yang2020fda, wang2021domain}. While such approaches have yielded promising results, there has been a notable imbalance between the attention given to domain adaptation methods and the integration of foundational training practices. Many existing works tend to emphasize domain adaptation techniques while neglecting well-established best practices derived from training on real data. These practices encompass designing a robust augmentation pipeline, optimization and regularization of hyperparameters, tailored to the downstream task, and ensuring data diversity. All of these practices have proven instrumental in enhancing the generalization capability of object detection models \cite{he2019bag, steiner2021augreg}.

Another crucial aspect of object detection model design pertains to the choice of neural network architecture. CNNs have become the de facto standard backbone for various computer vision tasks, including object detection trained on synthetic data \cite{shermeyer2021rareplanes, kiefer2022leveraging, Vanherle_2022_BMVC}. However, over the past years research has shed light on a notable bias exhibited by CNNs. Empirical observations and theoretical analyses suggest that CNNs behave akin to high-pass filters, emphasizing texture-related features over shape-based features \cite{geirhos2018}. While these properties can be controlled for through e.g. augmentations \cite{geirhos2018}, without tailored interventions they can significantly impact model generalization. In contrast, Transformer architectures exhibit a distinct `shape bias' behavior, akin to a low-pass filter \cite{naseer2021intriguing, park2022how}. In the context of object detection, the texture bias of CNNs may hinder their ability to generalize effectively when trained on synthetic data, as synthetic data often struggles to faithfully capture fine-grained textures present in real-world scenes. Conversely, the shape bias inherent to Transformer architectures aligns more closely with the shared characteristics of synthetic and real data. Synthetic datasets tend to excel in accurately representing geometric shapes and structural layouts. Thus, the use of synthetic datasets as training data for a Transformer backbone is a potentially powerful combination.

This research paper aims to establish a strong baseline approach for training object detection models on synthetic data, through exploring the potential of leveraging shape bias in Transformer architectures and integrating traditional best practices for training object detection models, such as including strong data augmentation techniques.


\section{RELATED WORK}
\label{sec:relwork}

A scan of the state of the art has been conducted to gain a deeper understanding of methods for object detection, and the use of synthetic data and data augmentation for improved model performance. 

\subsection{Object detection}
\label{sec:object_detection}

Deep Learning (DL) \cite{lecun2015deep} has been a game-changer for object detection, achieving state-of-the-art results on a variety of benchmarks and applications, such as autonomous driving \cite{balasubramaniam2022object}, people counting \cite{krishnachaithanya2023people}, search and rescue operations \cite{domozi2020real}, and military object detection \cite{kong2022yolo}. DL-based object detectors typically employ CNNs to extract features from images. Most of these follow a two-stage \cite{cai2018cascade, girshick2015fast, ren2015faster, he2017mask} or a single-stage \cite{redmon2016you, redmon2017yolo9000, redmon2018yolov3, bochkovskiy2020yolov4, wang2023yolov7} object detection approach, adapted to the number of steps needed to solve the task. Attention to multi-scale features is often fundamental, and has led to developments from Feature Pyramid Networks (FPN) \cite{lin2017feature} to more elaborate versions such as the BiFPN component of EfficientDet \cite{tan2020efficientdet}. Recent advances show that the adoption of different architectures such as the computer vision adaptation of Transformers \cite{dosovitskiy2020image, wolf2020transformers} may result in improvements to the state-of-the-art results. An example is ViT-Yolo \cite{zhang2021vit}, which is a hybrid model combining convolution with the self-attention mechanism. DETR \cite{carion2020end} also exploits an encoder-decoder structure to achieve end-to-end detection. Lastly, the Swin Transformer \cite{liu2021swin} has proven to be a reliable backbone for several computer vision tasks. Due to the distinct `shape bias' behaviour that Transformers exhibit \cite{naseer2021intriguing, park2022how}, Transformer-based methods are arguably more robust to occlusion, perturbations and domain shifts compared to CNNs, and are more suitable when it comes to giving importance to the object's shape and geometrical features rather than its texture \cite{naseer2021intriguing}. This `shape bias' is likely caused by Transformers having a very high receptive field early in the network, allowing the formation of long-range connections early, while CNNs start with a local receptive field that grows gradually each layer \cite{raghu2022vision}.

\subsection{Synthetic data}
\label{sec:synthetic_data}

Synthetic data has been adopted as a valuable resource to address a lack of large quantities of annotated real-world training data, saving time in acquisition and manual annotation. Its use has been shown to improve the performance of object detection models in a number of different studies \cite{vanherle2022analysis, Eker2023, rajpura2017object, prakash2019structured}. In the context of computer vision, synthetic data can be defined as images or videos (i) generated from scratch, for example through computer graphics or rendering engines, or generative models like GANs or diffusion models \cite{azizi2023synthetic}, (ii) modified versions of existing real data, for example employing DL-based augmentations \cite{trabucco2023effective} or style transfer \cite{ho2020retinagan}, or (iii) a combination of the two, for example inpainting \cite{he2024image}. Besides saving time on data gathering and annotation, synthetic data can also offer the advantage of a large degree of precision and control over the environment through which synthetic datasets are built \cite{Eker2023}. However, synthetic data also comes with its challenges. Bridging the reality gap, as well as building datasets with enough variability to represent all possible target scenarios and characteristics sufficiently, can be a challenge in creating and using synthetic data. To avoid performance degradation caused by the reality gap, recent literature explores data diversity \cite{wang2019learning}, coherent in-context placement \cite{georgakis2017synthesizing}, coverage in viewpoint and scale distributions \cite{dwibedi2017cut}, variation in clutter and occlusions \cite{rajpura2017object}, unrealistic \cite{tremblay2018deep} or structured \cite{prakash2019structured} domain randomization \cite{tobin2017domain, borrego2018applying}, photorealism \cite{movshovitz2016useful, wang2021real} or a combination of the last two \cite{tremblay2018deep}.
 
\subsection{Data augmentation}
\label{sec:lit_augmentations}

One of the most effective ways to improve detection performance is the use of data augmentation techniques during training. However, we find that related work often retains the default augmentations provided by their framework or baseline model of choice, occasionally supplemented by photometric distortions such as color jitter and blurring. Certain object detection frameworks such as YOLOv5 \cite{glenn_jocher_2022_6222936} are optimized for fine-tuning on relatively small amounts of task-specific data, and thus have a strong augmentation pipeline, and are  continuously updated to follow new state-of-the-art approaches. For the Transformer backbone, we find that most common implementations such as those in MMDetection \cite{mmdetection} and Detectron2 \cite{wu2019detectron2} have a default set of augmentations optimized for large-scale pretraining, usually limited to random flips and some resizing and cropping. This stands in contrast with the recent advances in image augmentation, which have proven that model performance may increasing significantly when artificially increasing the size and diversity of a dataset, creating new images from existing ones \cite{yang2020fda}. On top of well-known ‘traditional’ augmentations (flipping, cropping, rotating, scaling, blurring), one can work on specific color space and intensity transformations, photometric distortions, blending, or use methods like AutoAugment \cite{cubuk2019autoaugment}, MixUp \cite{zhang2017mixup}, RandAugment \cite{cubuk2020randaugment}, CutMix \cite{yun2019cutmix}, or GAN-based augmentations \cite{frid2018synthetic}. These methods have shown to improve performance on object detection tasks \cite{cubuk2019autoaugment,zhang2017mixup, cubuk2020randaugment,yun2019cutmix,frid2018synthetic}. This is especially important when used in conjunction with a Transformer backbone \cite{steiner2021augreg}, as the weaker inductive biases of this backbone cause an increased reliance on regularization or augmentation, compared to CNNs. Due to the large search space of possible augmentations and their hyperparameters, it is computationally infeasible to do a full sweep over all possible permutations. RandAugment \cite{cubuk2020randaugment} is an automatic augmentation optimization scheme that is able to reduce the search space while automatically optimizing the applied augmentations with respect to validation accuracy.

\section{METHODOLOGY}
\label{sec:method}

This section highlights the methods used in our experiments in terms of backbone architecture and data augmentation methods, and elaborates on the selected datasets and their quality. 

\subsection{Backbone Architecture}
\label{sec:backbone}

As highlighted in the Introduction, CNNs behave similar to high-pass filters, emphasizing high-frequency texture-related features \cite{geirhos2018} while Transformers behave similar to low-pass filters, exhibiting a distinct bias to detect low-frequency shape-based features \cite{naseer2021intriguing, park2022how, raghu2022vision}. This is especially relevant in the domain of synthetic data, where it is often much easier to faithfully capture the shape of an object than it is to implement photo-realistic textures and lighting. We propose to leverage the aforementioned properties of the Transformer architecture. Thus, we employ several pre-trained Transformers and compare their performance when trained on synthetic data and evaluated on real data, to the performance of baseline convolutional models. The Swin Transformer \cite{liu2021swin} was selected to operationalize the Transformer backbone. This popular backbone is characterized by window-based self-attention: it performs local attention within a window. Additionally, Swin reaches similar or better performance on object detection tasks compared to other, more computationally expensive, Transformers \cite{liu2021swin,han2022survey}. The Swin Transformer does exhibit a significantly smaller shape bias than a full attention vision Transformer \cite{morrison2021exploring}, but that is considered to be a reasonable trade-off because the full attention computational costs remain prohibitive for the dense prediction tasks. For fair comparison with the baseline convolutional model, the size of the Swin Transformer was selected to match the parameter count of the baseline convolutional model as closely as possible.

\subsection{Data}
\label{sec:dataqual}
Data quality is crucial in training Deep Learning models. We find that many synthetic object detection datasets are limited in terms of variety in the 3D models used, the quality of their textures, and environmental factors such as lighting and backgrounds. Besides variation, we find that label quality is also a common concern. A common way to generate object detection labels is to draw a bounding box around every part of the target object that is visible from the camera viewpoint. While proper implementation should result in perfect ground truth labels, a manual inspection of the data revealed that mistakes were still present in the datasets. For example, DGTA-VisDrone \cite{kiefer2022leveraging} contains objects that are completely obscured by trees or foliage, presumably due to raycasts being used for calculating visibility ignoring those surfaces even though they are opaque. We perform experiments on the challenging VisDrone dataset \cite{Cao_2021_ICCV} with DGTA-VisDrone \cite{kiefer2022leveraging} as a synthetic training counterpart, the synthetic-to-real RarePlanes dataset\cite{shermeyer2021rareplanes}, and a synthetic and real in-house vehicle detection dataset from concurrent work \cite{Eker2023}. The details of each dataset are listed below. 


\subsubsection{VisDrone}
\label{sec:visdrone}

\begin{figure}[H]
\centering
\begin{subfigure}{.5\textwidth}
  \centering
  \includegraphics[width=.9\linewidth]{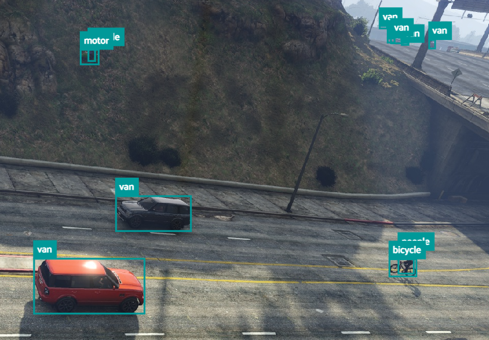}
  \caption{An example image of the DGTA-Visdrone dataset \cite{kiefer2022leveraging}.}
  \label{fig:VisDroneExampleTrain}
\end{subfigure}%
\begin{subfigure}{.5\textwidth}
  \centering
  \includegraphics[width=.9\linewidth]{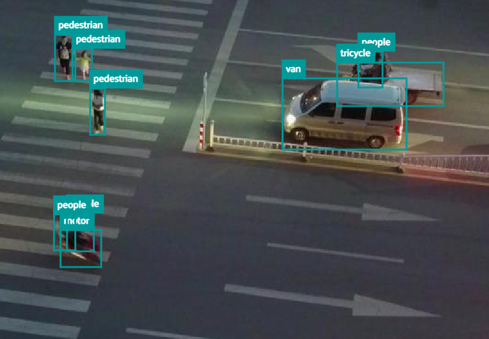}
  \caption{An example image of the VisDrone dataset \cite{zhu2018vision}.}
  \label{fig:VisDroneExampleTest}
\end{subfigure}
\caption{Example images for the synthetic/train dataset (a) and the real-world/validation dataset (b)}
\label{fig:VisDroneExample}
\end{figure}

The VisDrone dataset is a large-scale visual object detection and tracking benchmark \cite{zhu2018vision}. The dataset was captured using drones over various Chinese urban and suburban areas and consists of 263 videos and 10,209 images. The dataset was recorded over a variety of scenes across 14 cities in China, across various weather and lighting conditions. The dataset also contains a wide variety of viewpoints, including oblique angles and overhead viewpoints, from a range of low to high altitudes \cite{zhu2018vision}. The images are annotated using 10 fine-grained classes, including vehicles such as cars, vans, or busses, separate classes for pedestrians and drivers/cyclists, and fine-grained distinctions between bicycles, tricycles, and awning-tricycles \cite{zhu2018vision}. The dataset has been used in multiple challenges for evaluating detector performance on small objects \cite{Cao_2021_ICCV}. When trained on real data, the challenge has reached a performance of mAP 38-39\%, and mAP@50 of 62-65\% \cite{Cao_2021_ICCV}. As a counterpart to this real-world dataset, the synthetic DGTA-VisDrone dataset has been developed \cite{kiefer2022leveraging}. This dataset contains 50,000 images from drone viewpoints, with a great variety in altitude and angle. The dataset was created using the open-source framework DeepGTAV, based on the video game GTAV. The annotations of this dataset are limited to 8 out of 10 VisDrone classes, excluding the classes tricycles and awning-tricycles \cite{kiefer2022leveraging}. When training on this synthetic dataset, performance has thusfar reached a maximum mAP@50 of 10.2\%\cite{kiefer2022leveraging}. \autoref{fig:VisDroneExample} shows an image from the original VisDrone dataset alongside an example from the DGTA-VisDrone set. 

\subsubsection{RarePlanes}
\label{sec:rareplanes}
\begin{figure}[H]
\centering
\begin{subfigure}{.5\textwidth}
  \centering
  \includegraphics[width=.9\linewidth]{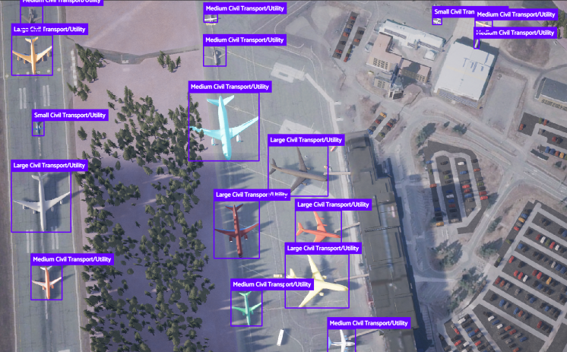}
  \caption{An example image for the RarePlanes dataset, synthetic portion \cite{shermeyer2021rareplanes}.}
  \label{fig:RarePlanesExampleTrain}
\end{subfigure}%
\begin{subfigure}{.5\textwidth}
  \centering
  \includegraphics[width=.9\linewidth]{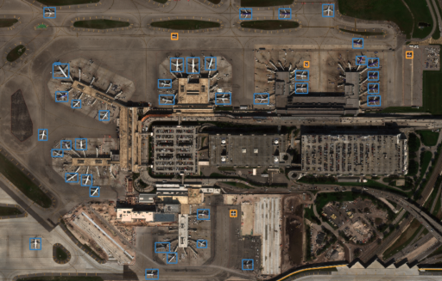}
  \caption{An example image of the RarePlanes dataset, real-world portion \cite{shermeyer2021rareplanes}.}
  \label{fig:RarePlanesExampleTest}
\end{subfigure}
\caption{Example images for the synthetic/train dataset (a) and the real-world/validation dataset (b).}
\label{fig:RarePlanesExample}
\end{figure}

The RarePlanes dataset is a collection of WorldView-3 satellite imagery capturing a wide variety of airplanes at 0.3m resolution \cite{shermeyer2021rareplanes}. The dataset includes both real and synthetic images. The real portion of the dataset consists of 253 satellite scenes taken over 112 locations with 14,700 annotations. The synthetic portion of the dataset consists of 50,000 synthetic satellite images with 630,000 aircraft annotations from a wide variety of locations. AI.Reverie’s simulation platform was used to capture the images at the same 0.3m resolution \cite{shermeyer2021rareplanes}. The RarePlanes datasets are annotated with 10 attributes rather than classes, such as the presence of canards, the propulsion system, and number of engines \cite{shermeyer2021rareplanes}. Related work has thus focused on classification by role \cite{shermeyer2021rareplanes, downes2023rareplanes}, specifically for the subset of civilian airplanes classified by their size (small, medium, or large). We  train our model on this classification task as well. Previous studies have reached a maximum performance of 75\% mAP and 93\% mAP@50 for this same task when training on real data \cite{downes2023rareplanes} and 35.9\% mAP and 59.1\% mAP@50 when training solely on synthetic data \cite{shermeyer2021rareplanes}. \autoref{fig:RarePlanesExample} shows one image from the real alongside one image of the synthetic portion of the RarePlanes dataset. 

\subsubsection{Vehicle Detection}
\label{sec:milvehicles}

\begin{figure}[H]
\centering
\begin{subfigure}{.5\textwidth}
  \centering
  \includegraphics[width=.9\linewidth]{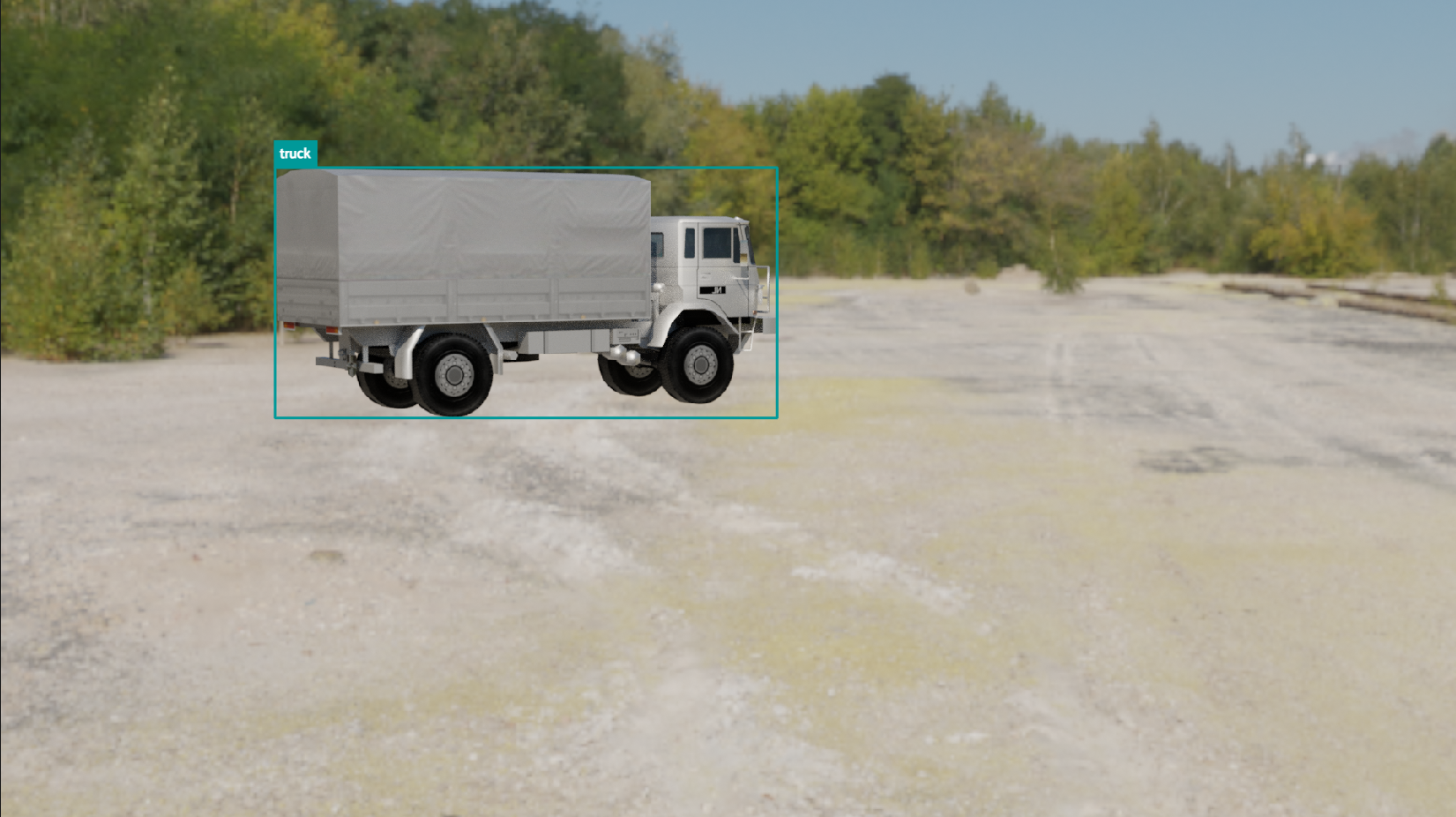}
  \caption{An example image of a truck from the vehicles dataset, synthetic portion \cite{Eker2023}.}
  \label{fig:MilVehExampleTrain}
\end{subfigure}%
\begin{subfigure}{.5\textwidth}
  \centering
  \includegraphics[width=.73\linewidth]{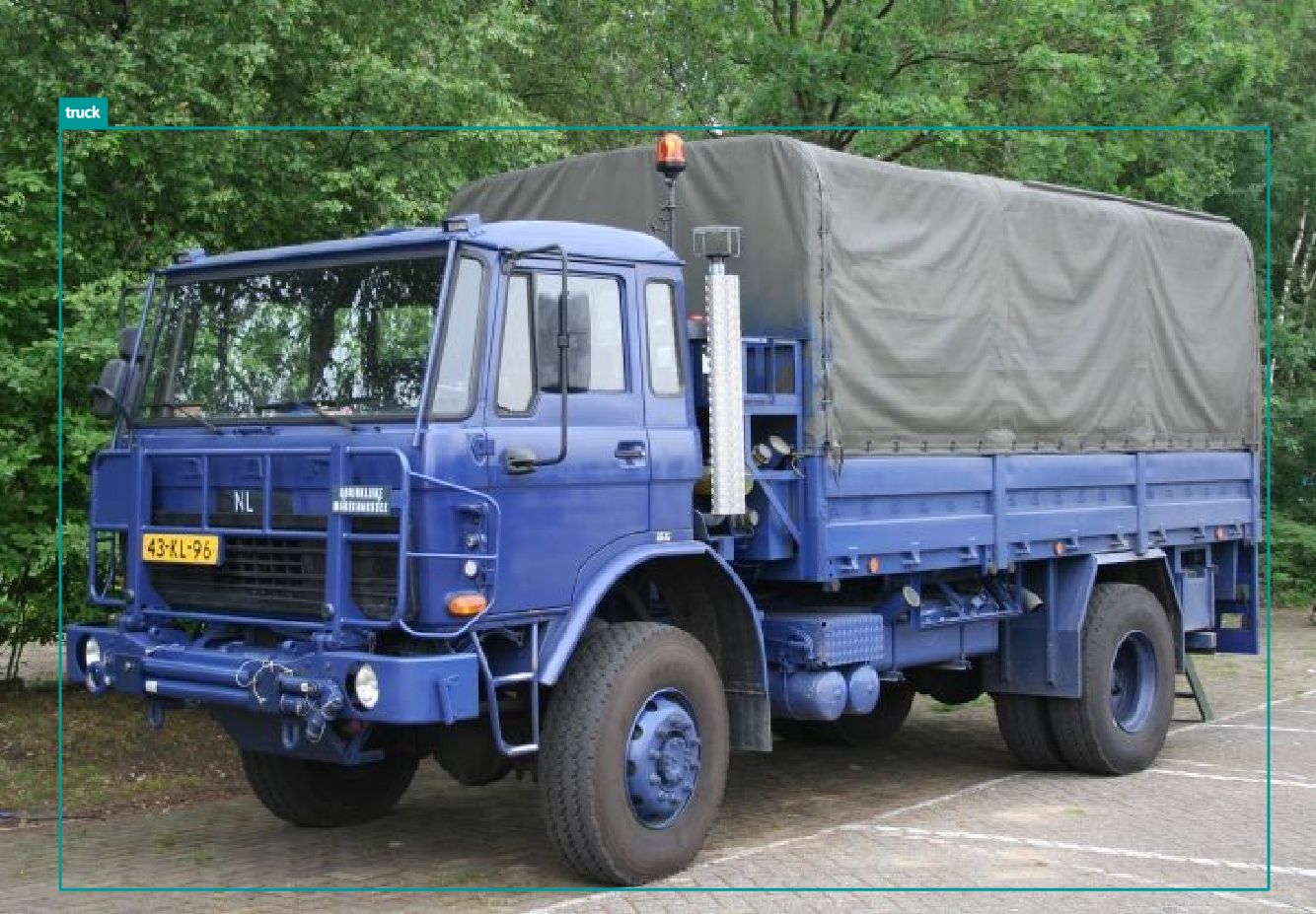}
  \caption{An example image of a truck from the vehicles dataset, real-world portion \cite{Eker2023}.}
  \label{fig:MilVehExampleTest}
\end{subfigure}
\caption{Example images for the synthetic/train dataset (a) and the real-world/validation dataset (b).}
\label{fig:MilVehExample}
\end{figure}

The last dataset we evaluated models on is one developed in-house for concurrent work \cite{Eker2023}. The dataset was designed through iterative variation of relevant axes, such as the number of images per object class, object-camera distance, and object pitch, to reach optimal mAP performance when training a vehicle object detector on synthetic data \cite{Eker2023}. The resulting dataset contains 1,600 synthetic and 333 real-world images of 4 types of vehicles. While the data construction method is relatively crude, consisting of just one 3D model per vehicle type superimposed on real photographs of various environments, the careful tuning of relevant simulation axes make it a very potent synthetic training set. The original publication managed to reach a performance of mAP 76.7\% and mAP@50 of 95.4\% when training on synthetic data and evaluating on real data \cite{Eker2023}. \autoref{fig:MilVehExample} shows one image from the real alongside one image of the synthetic portion of the vehicle detection dataset. 

\subsection{Augmentations}
\label{sec:augmentations}

\begin{figure}[H]
\centering
\begin{subfigure}{.5\textwidth}
  \centering
  \includegraphics[width=.9\linewidth]{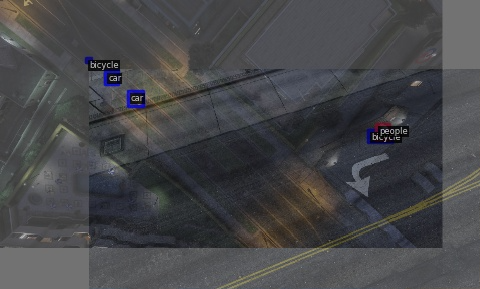}
  \caption{An example of the MixUp augmentation.}
  \label{fig:MixUp}
\end{subfigure}%
\begin{subfigure}{.5\textwidth}
  \centering
  \includegraphics[width=.73\linewidth]{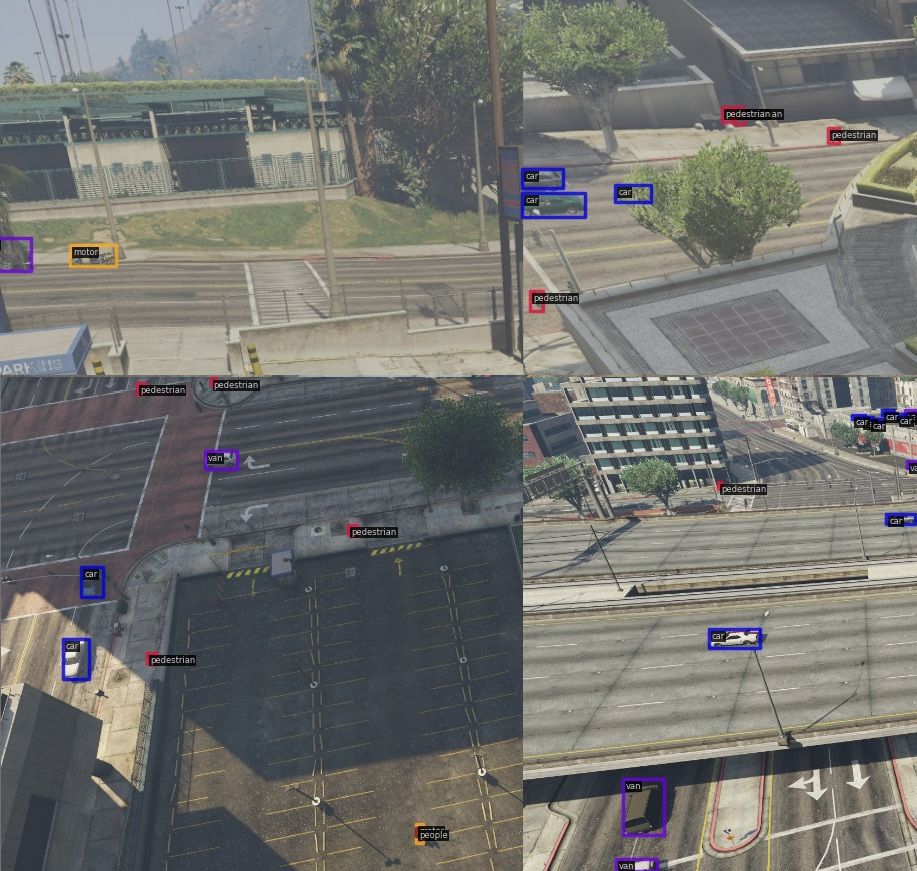}
  \caption{An example of the Mosaic augmentation.}
  \label{fig:Mosaic}
\end{subfigure}
\caption{Examples of the MixUp and Mosaic data augmentation methods, applied to the DGTA-VisDrone dataset.}
\label{fig:augmentations_examples}
\end{figure}

Unless stated otherwise, we will apply data mixing augmentation methods MixUp \cite{zhang2018mixup}, Mosaic \cite{bochkovskiy2020yolov4}, and large scale jittering \cite{ghiasi2021simple}. An example of MixUp and Mosaic is provided in Figure \ref{fig:augmentations_examples}. We will show that adopting a strong augmentation pipeline can significantly increase performance for the Transformer baseline. However, we show through the experiment described in section \ref{exp_rareplanes} that these improvements are not universal, and that the specific set of augmentations should always be carefully adjusted to the task at hand. We use RandAugment in our experiments to select a good augmentation policy from a set of photometric distortions: brightness, contrast, pixelation, jpeg compression, and gaussian blur.


\section{EXPERIMENTS}
\label{sec:experiments}

We perform experiments on three pairs of datasets of synthetic and real-world images. The baseline comparison models and their performance have been taken from the original publications in the case of DGTA-Visdrone \cite{kiefer2022leveraging} and RarePlanes \cite{shermeyer2021rareplanes}, or have been trained specifically for this publication if there was no baseline available, as was the case for the vehicle detection dataset \cite{Eker2023}. For fair comparison, the Swin Transformer \cite{liu2021swin} counterpart of the convolutional benchmark models were selected to have a roughly equal or lower parameter count. Our results are reported using either mAP, mAP@50, or both, based on the availability of comparative metrics in related work, as well as how representative these metrics are for the results of the specific experiment. In the following subsections we will detail the experimental setup for each individual dataset. 


\subsection{VisDrone}

For the VisDrone experiment, the (DGTA-)VisDrone dataset described in \autoref{sec:visdrone} was used. We compare to the ResNeXt-101 and YOLOv5 models from the original publication as benchmarks \cite{kiefer2022leveraging}. The ResNeXt-101 had been trained with limited data augmentations, while the YOLOv5-X model has a lower parameter count but more elaborate data augmentations. For our experiments we employed a Swin-S Transformer due to its similarity in size, with 66.1M parameters. The larger Swin-B would be a closer fit, but we did not find pretrained Faster-RCNN models utilizing this backbone. All data augmentations described in section \ref{sec:augmentations} were applied to our dataset for training. To evaluate whether a potential improvement in performance was caused by the data augmentations or the Transformer backbone, we trained a second Swin-S Transformer on the dataset without applying data augmentations. Finally, we train a DINO model with Swin-L backbone to see the impact of a much larger model on detection performance. We report all our results using mAP@50, due to the unavailability of mAP for the original publication. 


\subsection{RarePlanes}
\label{exp_rareplanes}
For the Rareplanes experiment, the RarePlanes dataset described  in \autoref{sec:rareplanes} was used. We compare to the benchmark ResNet-50 model from the original publication \cite{shermeyer2021rareplanes}, focusing on the object detection based on role, excluding military classes in accordance with the original publication \cite{downes2023rareplanes}. We performed the same object detection task with our model, which in essence is a classification based on size (small, medium, large) of civilian airplanes. This classification objective, combined with the fact that all imagery is taken at the same resolution of 0.3m per pixel and no cloud cover or other occlusions are present. This meant that most of the augmentations that were indispensable earlier, such as large scale jittering and MixUp, now harm detection accuracy. Therefore, we only used RandAugment in our best training run. We report our results using mAP, as this was the metric provided in the original publication.

\subsection{Vehicle Detection}
For the vehicle detection experiment, the vehicles dataset described in \autoref{sec:milvehicles} was used. There are no existing benchmarks since this is a newly published dataset developed in concurrent work \cite{Eker2023}. Therefore, we first trained a Transformer model which balanced efficiency in terms of parameter count with performance, and selected a convolutional model, ResNet-101, that had a similar parameter count as our final model. We additionally used a higher-performance backbone and detector architecture in the form of ResNeXt-101 64x4d and Cascade R-CNN to see if more compute may close the gap in model performance. All data augmentations were applied to the training data for both models. We report our results using both mAP and mAP@50. Due to the nature of our dataset, reporting solely the mAP would provide an inaccurate representation of the results, because a low mAP but high mAP@50 is observed to be the result of the exclusion of specific parts sticking out of the vehicle, such as an antenna, camera, or sometimes even the entire truck cabin. We do not consider the detection of these subparts to be a crucial feature of the object detector, and therefore provide both metrics but focus on the mAP@50 in our reporting. 




\section{RESULTS}
\label{sec:results}

This section provides the results of the experiments one by one, followed by a discussion of the results and suggestions for future research.

\subsection{VisDrone}

Table \ref{tab:VisDroneResults} shows the results of the VisDrone experiment. The results demonstrate improvements to the benchmark performance both through applying the Transformer architecture as well as enhancing augmentations. The benchmark ResNeXt-101 in particular had resulted in low performance, potentially caused by the lack of data augmentations in the orginal setup. The YOLOv5 benchmark already achieved better results, leveraging strong augmentation methods and thus preventing overfitting on synthetic data through domain randomization, to some extend \cite{kiefer2022leveraging}. However, the Transformer architecture, by leveraging a combination of data augmentation to prevent overfitting and leaning into the shape bias to make effective use of the consistency in shape from synthetic to real data, achieves significantly higher results with a lower parameter count. Our experiment with a combination of Transformer backbone and little data augmentations, demonstrated that proper augmentation remains crucial for high detection performance. Lastly, the experiment with the high-capacity DINO architecture and Swin-L backbone shows that even with suboptimal synthetic training data the model is able to reach impressive performance on the real VisDrone data. 

\begin{table}[H]
    \centering
    \begin{tabular}{c|c|c|c|c}
    \textbf{Backbone} & \textbf{Architecture} & \textbf{Parameters} & \textbf{Data augmentation} & \textbf{mAP@50} \\
    ResNeXt-101 64x4d$^*$ & Faster R-CNN & 127M & low &  2.4 \\
    CSP-Darknet53$^*$ &  YOLOv5-X & 86.7M & high &  10.2 \\
    Swin-S & Faster R-CNN & 66.1M & low & 7.8 \\
    Swin-S & Faster R-CNN & 66.1M & high &  16.2 \\
    Swin-L & DINO 5-scale & 218M & high &  26.1
    \end{tabular}
    \caption{The results of the VisDrone experiment, including benchmark results  \\ \footnotesize{$^*$ results taken from Kiefer et al. \cite{kiefer2022leveraging}.}}
    \label{tab:VisDroneResults}
\end{table}

\subsection{RarePlanes}

Table \ref{tab:RarePlanesResults} shows the results of the RarePlanes experiment. The results demonstrate a slight improvement to the baseline when using a Transformer backbone. As previously stated, a limited set of data augmentations was used for this experiment, thus we only evaluate the use of the Transformer backbone here. The Transformer backbone does show some improvement in mAP, but the performance gain is limited. The shape of an airplane is very distinctive, potentially working to the advantage of a Transformer backbone. However, the distinction between airplane classes is only based on size and therefore the shape bias does not offer as strong of an advantage here, leaving limited room for improvement by the Transformer backbone. Furthermore, even though the task seems relatively simple, the detection performance has been observed to be lowered predominantly due to the test dataset being tiled. This has resulted in a large number of planes for which just a small part of the wing or fuselage is visible. In such cases it is difficult to judge the actual size of the airplane, even for humans, and the coordinates of the bounding box corners are ambiguous due to the remainder of the plane not being visible. In addition to this, there is no occlusion or cloud cover in the real and the synthetic dataset, which are two other factors under which Transformers would be expected to perform relatively well.

\begin{table}[H]
    \centering
    \captionsetup{justification=centering,margin=2cm}
    \begin{tabular}{c|c|c|c|c}
    \textbf{Backbone} &  \textbf{Architecture} & \textbf{Parameters} & \textbf{Data augmentation} & \textbf{mAP} \\
    ResNet-50$^*$ & Faster R-CNN & 41.4M &  medium & 35.9 \\
    Swin-T &  Faster R-CNN & 45.2M & medium & 40.8
    \end{tabular}
    \caption{The results of the RarePlanes experiment, including benchmark results  \\ \footnotesize{$^*$ results taken from Shermeyer et al. \cite{shermeyer2021rareplanes}.}}
    \label{tab:RarePlanesResults}
\end{table}

\subsection{Vehicles Detection}

Table \ref{tab:MilVehResults} shows the results of the experiments with the in-house vehicle detection dataset. In this case, a strong improvement is demonstrated over the ResNet and ResNeXt architectures, reaching near-perfect object detection performance on the real dataset for a model trained solely with synthetic data. The high performance of the Transformer model could be the result of the strong emphasis on shape for classifying vehicles. This method focused strongly on shape and the images did not include depth or shadow, making them a particularly good fit for the Transfomer backbone and potentially harming their usefulness for a convolutional backbone. Even a larger convolutional model, the ResNeXt-101 64x4d, was not able to reach a similar performance to the Transformer, which was almost half its size in terms of parameter count. 


\begin{table}[H]
    \centering
    \begin{tabular}{c|c|c|c|c|c}
    \textbf{Backbone} & \textbf{Architecture} & \textbf{Parameters} & \textbf{Data augmentation} & \textbf{mAP@50} & \textbf{mAP}\\
    ResNet-101 & Faster R-CNN & 60.4M & high &  78.0 &  55.0 \\
    ResNeXt-101 64x4d & Cascade R-CNN & 127M & high &  87.5 &  66.0 \\
    Swin-S & Faster R-CNN & 66.1M & high & 94.8 & 80.1 \\
    Swin-T$^*$ & Faster R-CNN & 45.2M & high & 95.4 & -
    \end{tabular}
    \caption{The results of the vehicle detection experiment. \\ \footnotesize{$^*$ result taken from Eker et al. \cite{Eker2023}}}
    \label{tab:MilVehResults}
\end{table}

\section{Discussion}

This study demonstrated the capabilities of the Transformer architecture as well as data augmentation techniques to significantly improve object detectors trained on synthetic data. The VisDrone experiment showed how the existing benchmark performance could be improved both through data augmentations and a Transformer backbone, individually and combined. The performance gain could be explained by the strong shape bias of Transformer backbones, versus the texture bias of convolutional backbones. This causes the network to perform well in tasks that are heavily focused on the object shape, such as the detection of persons and (different types of) vehicles. The RarePlanes experiment showed that, while strong augmentations and choice of backbone are important, they need to be well-aligned with the task at hand. Notably, augmentations that were crucial for good performance on the VisDrone dataset had a negative effect on the RarePlanes dataset. Lastly, we presented object detection on a vehicle detection dataset developed in concurrent work. The Transformer backbone far outperformed the convolutional backbone for this task, even when using a convolutional backbone with nearly twice as many parameters. The Transformer backbone's shape bias is expected to have had a a strong impact on this result in particular due to the crude data construction method which featured 3D models that were limited both in quantity and quality. Nevertheless, the near-perfect performance on the real dataset with a model trained solely on synthetic data is impressive. 

In conclusion, our work demonstrates that a Transformer architecture can provide a potent backbone when training on synthetic data, potentially prior to domain transfer to achieve optimal results. The Transformer backbone's inherent shape bias makes it useful for object detection tasks that are heavily focused on shape, and less for object detection tasks with ambiguous shapes or tasks that are  heavily focused on texture. In addition to this, data augmentation methods are useful for object detection focused on shape as well, though they should be carefully selected based on the specifics of the dataset. Both these methods contribute to our suggestion to improve object detectors trained with synthetic data by starting with a stronger baseline methodology. 




\subsection{Limitations}

Several limitations of the research should be taken into consideration. First of all, the dataset pairs of synthetic and real data were limited in availability, size, and quality. The VisDrone dataset, though the largest dataset presented here, is limited in its synthetic counterpart, which lacks a proper class to class mapping and presents significant label noise. The RarePlanes dataset presents a relatively simple task, where the distinction between classes is made solely based on size with a consistent resolution and no cases of obstruction of the view or occlusion. There is also significant ambiguity in test set labels due to tiling of the images, resulting in many bounding boxes that only show a small part of a fuselage or wing. Lastly, the synthetic vehicle detection dataset was limited in terms of the quality of the original 3D models used. Besides these limitations in the existing datasets, there also exists a lack of high-quality object detection benchmark models adapted to these and other datasets. This makes it difficult to make fair comparisons between baseline and proposed methodologies. Given the existing benchmark models we have tried to create fair comparisons through using similar parameter count and presenting even larger convolutional models when it added value.  




\subsection{Future work}

We identified several potential courses for future research. The creation of high-quality and large synthetic and real dataset pairs would be a valuable contribution to this field. Improvement in quality could be made both through ensuring proper class-to-class mapping, improving the annotation quality, and improving the richness of context through using custom simulations. Additionally, the data construction method for synthetic datasets provides for interesting topics of research. There is a clear distinction between datasets that are based on the simulation of a single 3D object, such as the vehicle detection dataset, versus datasets that represent an entire 3D world, such as the DGTA-VisDrone and RarePlanes datasets. Leaning into this distinction and its impact on the usefulness of synthetic data for training could also be a topic of further investigation. 

The creation of more and stronger benchmark models for synthetic to real data would be a valuable addition to this field. The lack of such models limited the ability to make fair comparisons in this paper, and thus hinders the development of good practices for training on synthetic data. In this study, we focus on a type of object detection problem for which shape is the most salient factor. However, there are cases imaginable where a texture bias would be more relevant, for example when segmenting water, sky, or other surfaces with no well-defined shape. For these situations there are methods available that, e.g., propose hybrid architectures \cite{park2022how}, or add-on modules to inject high-frequency features for Transformers \cite{vitmatte}. These methods present interesting topics for further study and we strongly recommend comparison between Transformer, convolutional, and hybrid architectures to gain understanding of the most potent fits between backbones and object detection tasks. In particular, recent progress in applying full attention Transformers to dense prediction tasks \cite{chen2022vision, li2022exploring} may improve results even further due to their superior shape bias over local attention Transformers. 

\bibliography{report} 
\bibliographystyle{spiebib} 

\end{document}